%% file: 0_Main.tex
\documentclass[sigconf]{acmart}
\usepackage{latexsym}
\usepackage{array}
\usepackage{multicol}
\usepackage{multirow}
\usepackage{tabu}
\usepackage{caption}
\usepackage{subcaption}
\usepackage{bm}
\usepackage{CJK}
\usepackage{amsmath}
\usepackage{threeparttable}
\usepackage{lineno,hyperref}
\usepackage{algorithm}
\usepackage{algorithmic}
\usepackage{enumitem}
\usepackage{xspace}
\usepackage{fancyhdr}
\AtBeginDocument{%
  \providecommand\BibTeX{{%
    \normalfont B\kern-0.5em{\scshape i\kern-0.25em b}\kern-0.8em\TeX}}}

\setcopyright{acmlicensed}
\copyrightyear{2021}
\acmYear{2021}

\acmConference[CIKM '21] {Proceedings of the 30th ACM International Conference on Information and Knowledge Management}{November 1--5, 2021}{Virtual Event, Australia.}
\acmBooktitle{Proceedings of the 30th ACM Int'l Conf. on Information and Knowledge Management (CIKM '21), November 1--5, 2021, Virtual Event, Australia}
\acmPrice{15.00}
\acmISBN{978-1-4503-8446-9/21/11}
\acmDOI{10.1145/3459637.3482407}

\begin{document}
\fancyhead{}



\title{A Neural Conversation Generation Model via Equivalent Shared Memory Investigation}









\author{Changzhen Ji$^1$, Yating Zhang$^2$, Xiaozhong Liu$^3$, \\
Adam Jatowt$^4$,
Changlong Sun$^2$, Conghui Zhu$^1$ and Tiejun Zhao$^1$}
\affiliation{
\institution{
$^1$Harbin Institute of Technology, Harbin, China\\
$^2$Alibaba Group, Hangzhou, China\\
$^3$Worcester Polytechnic Institute, Worcester, Massachusetts, USA\\
$^4$University of Innsbruck, Innsbruck, Austria
}}
\affiliation{
\institution{
\small\tt 
czji\_hit@outlook.com, 
ranran.zyt@alibaba-inc.com,
liu237@indiana.edu, \\
adam.jatowt@uibk.ac.at,
changlong.scl@taobao.com, 
\{conghui,tjzhao\}@hit.edu.cn
}}


\begin{abstract}
Conversation generation as a challenging task in Natural Language Generation (NLG) has been increasingly attracting attention over the last years. A number of recent works adopted sequence-to-sequence structures along with external knowledge, which successfully enhanced the quality of generated conversations. Nevertheless, few works utilized the knowledge extracted from similar conversations for utterance generation. 
Taking conversations in customer service and court debate domains as examples, it is evident that essential entities/phrases, as well as their associated logic and inter-relationships can be extracted and borrowed from similar conversation instances. 
Such information could provide useful signals for improving conversation generation. 
In this paper, we propose a novel reading and memory framework called Deep Reading Memory Network (DRMN) which is capable of remembering useful information of similar conversations for improving utterance generation.
We apply our model to two large-scale conversation datasets of justice and e-commerce fields.
Experiments prove that the proposed model outperforms the state-of-the-art approaches.
\end{abstract}





\maketitle
\input{1_Introduction}
\input{2_Related_work}
\input{3_Model}

\input{4_Experiment}

\input{5_Result_discussion}

\input{6_Conclusion_outlook}
\bibliography{acmart}
\bibliographystyle{ACM-Reference-Format}










\end{document}

%% file: 1_Introduction.tex
\vspace{-12pt}
\section{Introduction}
\begin{figure*}
    \centering
    \includegraphics[width=\textwidth]{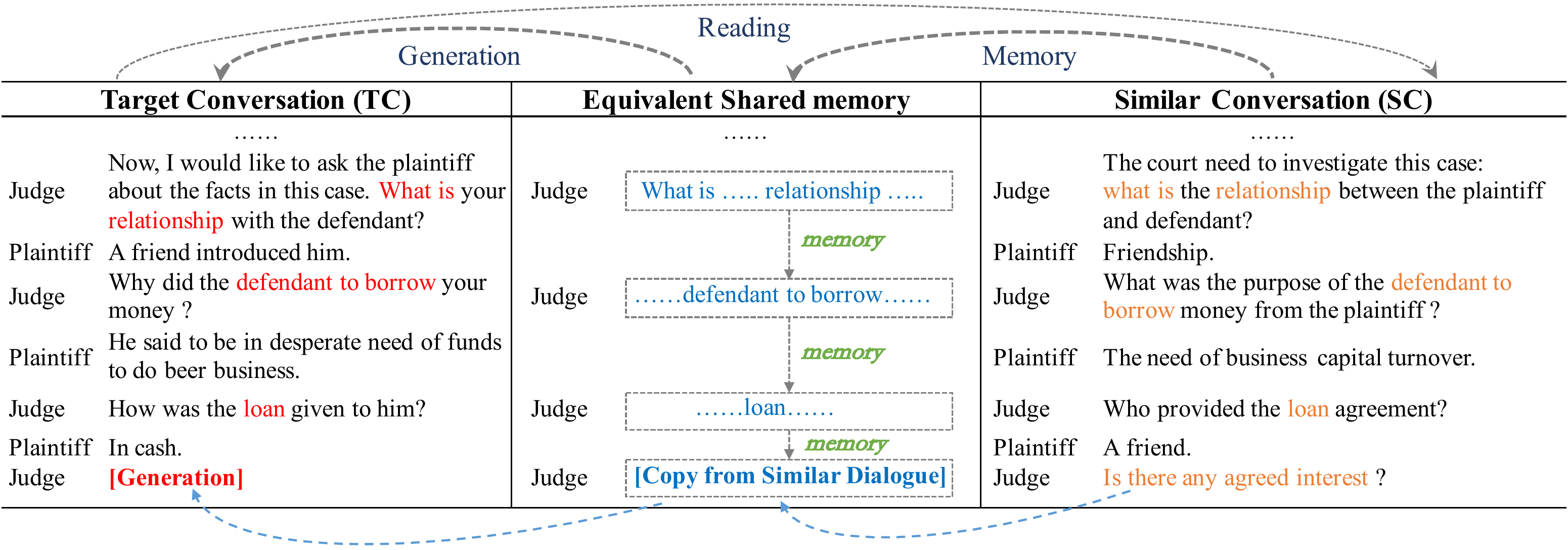}
    \vspace{-14pt}
    \caption{A toy case in our judicial dataset. We can generate the next utterance by reading similar conversations and memorizing related entities, phrases as well as sentences.}
    \vspace{-8pt}
    \label{fig:toy}
\end{figure*}

Over the past years, chatbot technologies and conversation mining approaches have been actively explored and applied in many tasks and applications for a variety of purposes including supporting users in their decision making, e.g., in e-commerce customer service and legal justice consulting. These groundbreaking developments typically utilize big data and deep learning technologies, which harness in-depth semantic and discourse information of conversations and apply sophisticated learning models.

In contrast to classical rule-based \cite{10.1145/365153.365168} and template-based \cite{williams2007partially, schatzmann2006survey} approaches applied for conversation generation, sequence-to-sequence models \cite{sutskever2014sequence,luong2015effective,bahdanau2014neural} are able to understand sequential dependencies between conversation utterances which is crucial for content generation \cite{shang2015neural,vinyals2015neural}.
Recently, external knowledge has been also utilized to further improve the performance of utterance generation. For instance, open-domain, unstructured knowledge has been employed \cite{xia2017deliberation, ye2020knowledge, ghazvininejad2018knowledge, parthasarathi2018extending}, while in other cases, domain-specific knowledge bases (organized based on triples) have been used in task-oriented conversations \cite{wu2019proactive,liu2018knowledge,madotto2018mem2seq,dhingra2016towards}. Furthermore, large-scale knowledge graphs have also recently been employed \cite{he2017learning,moon2019opendialkg,xu2020knowledge}.
Although the above-mentioned works allowed achieving superior results, constructing external knowledge bases can be quite an expensive and arduous task. Additionally, the low adaptability and transferability of domain knowledge restrict their real-world applications. Therefore, further efforts are required in order to address these challenges.

Equivalent Shared Memory (ESM) is a phenomenon that can be observed in conversation corpora, especially, between conversations belonging to the same domain. 
While the detailed contents of different conversations vary, there are certain common patterns which can be considered as the common backbone memory.
Such shared memory can be then useful for supporting the utterance generation task.
As portrayed in the example in Figure \ref{fig:toy}, a target legal conversation discourse (one on the left) and a selected reference prior conversation (on the right) share certain common patterns (i.e., ESM). 
Then the utterance generator is able to copy the utterance (in this case the judge's question) from the reference conversation and paste it into the Target Conversation (TC) context. This scenario can also occur in other conversation corpora, e.g., customer service, where an agent may repeat the same or a similar responses as ones in past dialogues.

Motivated by these observations, we propose a novel model to discover ESM by extracting critical words, phrases, and discourse information from similar conversations. The proposed model, the Deep Reading Memory Network \textbf{(DRMN)}, tries to reproduce the human decision-making process, in resemblance to a human brain which "retrieves" similar memories to utilize them for generating the next utterance in an ongoing conversation. 
\textbf{DRMN} treats the last utterance\footnote{The last utterance is usually most relevant to the forthcoming utterance to be generated.} \cite{zhang2018modeling}
as a query and Similar Conversations (SC) as a search database. The query is issued to retrieve from the search database the information most relevant for the next utterance to be issued.
 
To verify this hypothesis and to validate the proposed model, we conduct experiments on two large-scale conversation datasets from different domains - \textit{court debate dataset} from a legal field and \textit{customer service dataset} from an e-commerce field. We apply \textbf{DRMN} to the above two datasets and assess the model performance based on both automated and human evaluation. The experimental results indicate that \textbf{DRMN} significantly outperforms the baseline models.
 
Our study contributes to the growing body of research in exploring conversation generation. The particular contributions of this paper are as follows:
 
\begin{itemize}[leftmargin=*,topsep=2pt]
\item We propose a novel end-to-end model, called the Deep Reading Memory Network, for the conversation generation task to explore Equivalent Shared Memory from past conversations. 

\item We demonstrate that the proposed model has sufficient domain adaptability and generalization ability by experimenting on two datasets that originate from different domains - \textit{court debate} and \textit{customer service} conversation datasets. Experimental results show that our model produces the best results on both the datasets compared to the state-of-the-art methods.

\item To support and motivate other scholars for further investigating this novel and an important research problem, we make the experimental datasets and code publicly available\footnote{https://github.com/jichangzhen/DRMN}.
\end{itemize}

\begin{figure*}
    \centering
    \includegraphics[width=.95\textwidth]{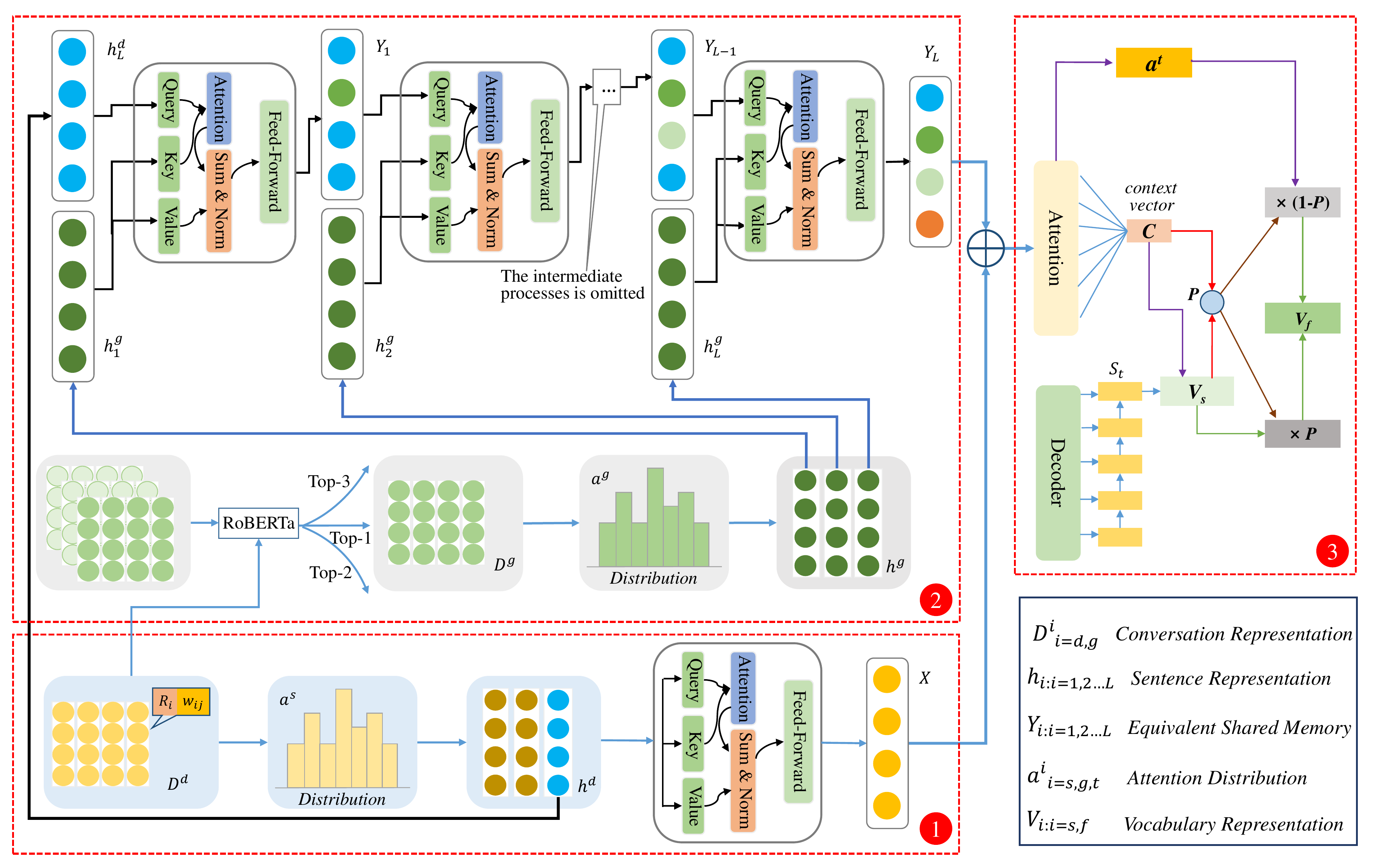}
    \vspace{-12pt}
    \caption{The overall structure of \textbf{DRMN}. The model is divided into three components, (1) Conversation Representation: it is used for encoding the target conversation; (2) Equivalent Shared Memory: it is used to retrieve the similar conversations and construct ESM and (3) Utterance Generation: it is used for generating utterances.}
    \vspace{-6pt}
    \label{fig:network}
\end{figure*}

%% file: 2_Related_work.tex
\vspace{-12pt}
\section{Related work}
\subsection{Conversation Generation}
Maintaining intelligent conversations to facilitate life in the real world has been the long-term goal of Artificial Intelligence (AI). In this regard, recently, the research in conversation generation, which is an important task in Natural Language Processing (NLP), has generated impressive achievements.

In the early years, researchers mostly adopted rule-based and template-based methods: 
for example, Joseph \textit{et al.} \cite{10.1145/365153.365168} proposed to generate responses by reassembling rules associated with selected decomposition rules.
Jost \textit{et al.} \cite{schatzmann2006survey} attempted at building systems that learn what constitutes a good conversation strategy through trial-and-error interaction.

In the recent years, due to the capabilities of deep neural networks, sequence-to-sequence models \cite{sutskever2014sequence,luong2015effective,bahdanau2014neural,hochreiter1997long,nallapati2016abstractive,kalchbrenner2016neural} become popular. They are widely used in conversation generation allowing to achieve significant improvements.
For example, \cite{shang2015neural,vinyals2015neural} supplement the classic attention model with contextual information.
\cite{zhao2017learning,duboosting,csaky2019improving} solve the problem of the lack of diversity and boring responses being generated for open-ended utterances.
The copy mechanism \cite{gu2016incorporating,see2017get} and hierarchical LSTM \cite{yang2016hierarchical} led to increased research in text generation. EED \cite{pandey2018exemplar} used example vectors to guide the generation of dialogue. CCN \cite{ji-etal-2020-cross} applied a hierarchical encoder and cross-copying method to the field of conversation generation. 
The introduction of Transformers \cite{vaswani2017attention} brought text generation to a new level. Tranformer-based models are also widely used in conversation generation \cite{zhang2019recosa, cai2019retrieval,zhang2018modeling}.

More recently, employing external knowledge to conversation generation became a popular approach: 
\cite{xia2017deliberation, ye2020knowledge, ghazvininejad2018knowledge, parthasarathi2018extending} utilized unstructured knowledge for conversation generation. 
Additionally, structured knowledge triples  have also been widely employed in the conversation generation task \cite{williams2017hybrid,wu2019proactive}.
Knowledge graphs as larger-scale external knowledge sources can also be utilized in conversation generation tasks \cite{zhang2019grounded,moon2019opendialkg,xu2020knowledge}.

Existing works have achieved superior results, but the extensive external knowledge construction and difficulty of domain adaptation restrict their real-life applications.
We note that the model proposed in this paper does not use any external knowledge and can achieve good results in the general-domain conversation generation.
\vspace{-12pt}
\subsection{Memory Networks}
Sequence-to-sequence models rely on RNN \cite{elman1990finding} and LSTM \cite{hochreiter1997long} to improve the word-dependency memory in the sentence. 
However, the memory capacity of RNN \cite{elman1990finding} and LSTM \cite{hochreiter1997long} is very limited. They usually only remember a dozen time steps at most. 
Therefore, when the length of utterance increases or the number of utterances in a dialogue grows, the sequence-to-sequence models cannot satisfy the requirements of conversation generation systems.

Memory network \cite{weston2014memory} was originally proposed by Facebook AI. 
It was initially used for reasoning in question answering systems. 
Later, the end-to-end memory network was proposed \cite{weston2014memory,sukhbaatar2015end} to solve the problem of the memory being too short in traditional sequence-to-sequence models. Afterwards, \cite{grefenstette2015learning} proposed storage memory using stack and queue structure. \cite{miller2016key} introduced key-value memory networks, increasing the scale of memory based on an end-to-end memory network.

Memory networks have been also widely used in recent conversation systems:
\cite{kim-etal-2020-efficient} proposed a selectively overwriting mechanism for more efficient Dialogue State Tracking (DST) by the memory network;
\cite{tian-etal-2020-response} proposed to create the document memory with some anticipated responses.
\cite{lin-etal-2019-task} introduced a Heterogeneous Memory Networks (HMNs) to simultaneously utilize user utterances, conversation history, and background knowledge tuples.
\cite{chu-etal-2018-learning} proposed to use neural models to learn personal embeddings in conversation.
In addition, memory networks have been also widely used in other natural language processing tasks, for instance in:
text classification \cite{geng-etal-2020-dynamic, nguyen-etal-2020-relational},
question answering system \cite{khademi-2020-multimodal, han-etal-2019-episodic},
information extraction \cite{kolluru-etal-2020-imojie, saputra-etal-2018-keyphrases},
text generation \cite{lei-etal-2020-mart,maruf-haffari-2018-document} and
language models \cite{sap-etal-2020-recollection}.

Different from the previous memory network structures, \textbf{DRMN} uses the last utterance of the target conversation as a query, while every utterance in the equivalent shared memory is used as key and value. It also adopts self-attention structure in the memory module, and it allows multiple loops' filtering of effective information in the memory process.

%% file: 3_Model.tex
\vspace{-12pt}
\section{Deep Reading Memory Network}
In this section, we introduce the details of the proposed Deep Reading Memory Network \textbf{(DRMN)} model, in which we establish an ESM module between the context of the target conversation and other historical conversations which have similar semantics to help generate the next utterance. The overall framework of the model is shown in Figure \ref{fig:network}. The proposed framework has three major components:

\begin
{enumerate}[leftmargin=*]
\item \textbf{Conversation Representation}: 
We propose to encode each conversation fragment based on hierarchical infrastructure consisting of \emph{utterance representation layer} and \emph{conversation representation layer} (Section \ref{Conversation Representation}).

\item \textbf{Equivalent Shared Memory}: 
We introduce the method for discovering similar conversations by using the pre-trained RoBERTa model to retrieve candidates and then constructing Equivalent Shared Memory based on these candidates (Section \ref{Deep Reading Memory}).
 
\item \textbf{Utterance Generation}: 
Based on the extracted contextual information of the target conversation as well as the equivalent shared memory constructed from similar conversations, we further employ pointer generation mechanism to generate the next utterance (Section \ref{Utterance Generation}).
\end{enumerate}

\subsection{Conversation Representation}
\label{Conversation Representation}
In order to represent the conversation fragment, we make the following definition:
given a conversation $D=\{(U, R)^L\}$ containing $L$ utterances, $U$ and $R$ represent the utterance and the role of a speaker, respectively, where each utterance in the conversation is expressed as $U_i=\{w_{i1}, w_{i2},...,w_{il}\}$, with $w$ being a word and $l$ denoting the length of the utterance. Note that we use $D^d$ to represent target conversation, and the similar conversation will be denoted by $D^g$.

It should be pointed out that the role information can be critical for conversation generation since different characters taking part in the conversation may not necessarily share the same vocabulary space (e.g., plaintiff, defendant, customer or service staff).
Therefore, we take role information into consideration in conversation representation.
We concatenate the role $R_i$ with each utterance $U_i$ as a sentence: $S_i=[[R_i, w_{i1}], [R_i, w_{i2}],...,[R_i, w_{il}]]$.
The conversation can be expressed as $D = \{S_1,S_2,...,S_L\}$.
For utterance information, we utilize word2vec \cite{mikolov2013efficient} to construct the initial word vectors.
For role information, we take the randomly initialized vectors.

\subsubsection{Utterance layer}
The Bidirectional Long-Short Term Memory Network (Bi-LSTM) \cite{hochreiter1997long} has superior performance in representing sequential text. We apply it to hierarchically encode each conversation. 
For the utterance layer, the encoder represents a sentence with hidden representation $h^d_i = \{h^d_{i1},h^d_{i2},...,h^d_{il}\}$, as defined below:
\begin{equation}
\label{Eq:lstm}
    h^d_{ij} = \mathbf{Bi-LSTM}(e(S_{ij}), h^d_{ij-1})
\end{equation}
where $e(S_{ij})$ is the embedding of $S_{ij}$. $i$ represents the i-th utterance in the conversation, while $j$ represents the j-th word in the current utterance.

We then use the attention mechanism \cite{bahdanau2014neural} to estimate the importance of words in the sentence expressed as word-level attention distribution $a^s$: 
\begin{equation}
\label{Eq:hd}
h^d_i=\sum_{j=1}^{l}a_j^sh^d_{ij}
\end{equation}

\begin{equation}
\label{Eq:aj}
a_j^s = {\frac{exp(tanh(W^dh^d_{ij}+b^d)^{\mathrm{T}}h^d_{ij})} 
{\sum\nolimits_{j=1}^lexp(tanh(W^dh^d_{ij}+b^d)^{\mathrm{T}}h^d_{ij})}}
\end{equation}

\subsubsection{Conversation layer}
Similarly, for the conversation layer, in order to obtain the dependencies between sentences, we again use Bi-LSTM to encode $h^d_i$:
\begin{equation}
    h^d_{i} = \mathbf{Bi-LSTM}(e(S_{i}), h^d_{i-1})
\end{equation}

We also obtain the importance of different sentences in the conversation expressed as sentence-level attention distribution $a^u$: 
\begin{equation}
h^d=\sum_{j=1}^{l}a_i^uh^d_{i}
\end{equation}

\begin{equation}
a_i^u = {\frac{exp(tanh(W^uh^d_{i}+b^u)^{\mathrm{T}}h^d_{i})} 
{\sum\nolimits_{i=1}^lexp(tanh(W^uh^d_{i}+b^u)^{\mathrm{T}}h^d_{i})}}
\end{equation}

$W^d$, $b^d$, $W^u$ and $b^u$ are learnable parameters and $tanh$ is hyperbolic tangent function.
Thereby, we finally obtain the target conversation representation $h^d$.

\subsection{Equivalent Shared Memory}
\label{Deep Reading Memory}
As mentioned before, we are motivated by the observation of Equivalent Shared Memories (ESM) existing across different conversations. The proposed model is established based on
the hypothesis that ESM can be discovered from similar conversations and employed to generate the next utterance of the target conversation. 

\subsubsection{Similar Conversations Discovery}
Similar conversations play a key role in \textbf{DRMN}. 
First, we introduce how similar conversations are obtained.
Our goal is to find similar conversations to the target conversation.
Due to a typically large number of samples in real-world datasets, to assure high efficiency of retrieving similar conversations we use ElasticSearch\footnote{\url{https://www.elastic.co/products/elasticsearch}} to retrieve the top $50$ similar conversations as candidates by leveraging the target conversation as a query and the other samples as documents. 
To capture semantic information, we fine-tune the pre-trained RoBERTa model \cite{liu2019roberta}. Then, we add a dense layer with softmax as a classifier to obtain the semantic similarity score between the target conversation and each candidate. 

Next, we describe how the similar conversations are represented.
We obtain the representation of a similar conversation by using the same way as representing the target conversation, with the difference that the similar conversation fragment is encoded only with the utterance layer. This is because, to construct the ESM, the context of the target conversation interacts with each sentence in the similar conversations.
Then the similar conversation $D^g$ is represented in the way as illustrated in Eqs. \ref{Eq:lstm}, \ref{Eq:hd} and \ref{Eq:aj}. We then obtain the attention distribution $a^g$, as well as the representation of each sentence in the similar conversation, which is expressed as: $h^g = [h^g_1, h^g_2,..., h^g_L]$. 




\subsubsection{Equivalent Shared Memory Construction}
Equivalent Shared Memory (ESM) refers to the backbone patterns that commonly appear in similar conversations, which are closely related to the sentences to be generated.

Note that the sentences in the target conversation and ones in each selected similar conversation are not in one-to-one correspondence; for example, the utterance to be generated in the target conversation could be similar to the third or fifth utterance in the similar conversation, or to any other subset of sentences. So we need to read every utterance in the similar conversation to construct ESM.

Take a hypothetical judicial scenario as an example. The construction of ESM could be viewed as simulating the way in which the experience of a judge gradually grows through learning from similar cases. When a judge takes part in a trial case, he/she may first mentally go over the memories (or even physically check the related documents) to find similar conversations, and then to recall (or read) their entire trial process. In the end, he/she formulates his/her own utterances for the target conversation based on the common words/logic borrowed from the similar conversation fragments.

To explore the connection between the target conversation and the fragments of similar conversations,
we propose to let the last sentence appearing in the context of the target conversation interact with each utterance in the similar conversations (see procedure 2 depicted in Figure \ref{fig:network}). This is due to the observation that, in most cases in the conversation generation task, the utterance to be generated is more related to the last utterance in the context \cite{zhang2018modeling}.
The representation of the last sentence in the target conversation context is denoted as $h_L^d$ (using Equation \ref{Eq:hd}) while each sentence in the similar conversation is represented as $h^g = [h^g_1, h^g_2,..., h^g_L]$.

Unlike in the case of the traditional memory networks, to obtain the dependency between $h^d_L$ and $h^g_i$, the model reads $h^g_i$ in order, and at every step adopts the self-attention \cite{vaswani2017attention} module as memory structure. 
The self-attention can be expressed as:
\begin{equation}
    SA(\mathcal{Q}, \mathcal{K}, \mathcal{V}) =  softmax(\frac{\mathcal{Q} \cdot \mathcal{K}^T} {\sqrt{d}}) \cdot \mathcal{V}
\end{equation}
with three inputs: the query $\mathcal{Q}$, the key $\mathcal{K}$ and the value $\mathcal{V}$. This module first takes the query to attend to the key via Scaled Dot-Product Attention, then applies those attention results upon the value.

Inspired by the concept of self-attention, 
we use the last sentence $h_L^d$ as a query $\mathcal{Q}$, and each sentence in the similar conversation $h_i^g$ as the key $\mathcal{K}$ and value $\mathcal{V}$. The memory module is defined as follows:
\begin{equation}
SA(h^d_L,h_1^g,h_1^g) = softmax(\frac{{h^d_L} \cdot {h_1^g}^T} {\sqrt{d}}) \cdot h_1^g
\end{equation}
To prevent vanishing or exploding gradients, we adopt a layer normalization
operation \cite{ba2016layer} which refers to a feed-forward network $\mathcal{F}$ with RELU activation function \cite{goodfellow2016deep,zhou2018multi}:
\begin{equation}
\mathcal{F}(x) =  max(0; xW^f + b^f)W^h + b^h
\end{equation}
where $W^f$, $b^f$, $W^h$, and $b^h$ are learnable parameters.

We represent the memory in a time step $t$ as $Y_t$, and the first memory $Y_1$ can be represented as:
\begin{equation}
Y_1 = \mathcal{F}(SA(h^d_L,h_1^g,h_1^g))
\end{equation}

Our reading and memory update proceeds iteratively, and the previous memory is used as the input of the next memory.
The memory content will continue to be enriched as the amount of information increases.
Thus, this process can be described as:
\begin{equation}
  Y_t = \mathcal{F}(softmax(\frac{{Y_{t-1}} \cdot {h_t^g}^T} {\sqrt{d}}) \cdot h_t^g)
\end{equation}
since the length of the conversation is $L$. Finally, we can obtain the integrated memory which is denoted as $Y_L$.

\subsection{Utterance Generation}
\label{Utterance Generation}
To solve the long-dependency problem which often occurs in a multi-turn conversations, we apply self-attention mechanism to get the final representation of the target conversation, expressed as $X$. 
We merge the contextual distribution of the target conversation $X$, and the ESM discovered from the similar conversation $Y_L$ as the input for the decoder. 

During the decoding process, we selectively copy ESM.
This further solves the Out-Of-Vocabulary (OOV) \cite{see2017get} problem.
We learn a probability pointer $P$, which is used to determine whether to generate or copy at the current time step.

At the time step $t$, the decoder state, attention distribution, and context vector are represented as $s_t$, $a^t$, and $C$, respectively. The target vocabulary distribution is expressed as:
\begin{equation}
V_{s} = softmax(W^v(W^w[C^t,s_t]+b^w)+b^v)
\end{equation}

In addition, $P$ is computed as:
\begin{equation}
\label{Eq:alpa}
    P = \sigma(W^p(X+Y_L)+W^cC^t_t+W^ss_t+b^p)
\end{equation}

Therefore, we get the final vocabulary distribution $V_f$:
\begin{equation}
V_f = P*V_s + (1-P)*\sum_{i:w_{i}=w}^{l*L}a^t
\end{equation}

$W^w$, $b^w$, $W^v$, $b^v$,$W^p$ and $b^p$ in the three equations above are learnable parameters, $\sigma$ is sigmoid function and the $tanh$ is hyperbolic tangent function.

Finally, we use a cross-entropy loss as the objective function:
\begin{align*}
& loss = -\log P(U \mid D) \\
& \quad \  \, =-\sum_{j=1}^{l}\log P(w_{ij}\mid w_{i1:j-1},D)
\end{align*}

%% file: 4_Experiment.tex
\vspace{-8pt}
\section{Experimental Settings}
In this section, we introduce the datasets used for experiments, baselines, evaluation metrics, and training details.

\subsection{Datasets}
We conduct our experiments on both the court debate dataset in judicial field and the customer service dataset in e-commerce field to prove the effectiveness of our proposed model.
Both the datasets are constructed from real-world conversations. We have chosen the datasets belonging to quite different domains in order to prove that our model has good domain adaptability.

\subsubsection{Court Debate Dataset}
The Court Debate Dataset (CDD) is provided by the High People’s Court of one province in China. 
It contains $260,190$ trial multi-role conversations.
All the court transcripts are manually recorded by a legal professional.
In this work, we aim at generating the words spoken by the judge. Therefore, we consider the judge's utterance as the output of the model, and the previously-spoken context in the conversation is regarded as input.
The details of the dataset are shown in Table \ref{tab:table1}.

\begin{table}[!htbp]
\small
    \caption{Statistics of Court Debate Dataset}
    \label{tab:table1}
    \vspace{-5pt}
    \begin{threeparttable}
    \setlength{\tabcolsep}{4.5mm}
	\begin{tabular}{c|c|c|c}
	\hline
	 ~ & Conversation & Utterance & length(avg) \\
	 \hline
	Train & 208,152 & 2,869,794  & 7.2\\
	\hline
	Dev & 26,018  & 364,345 & 7.0\\
	\hline
	Test & 26,020  & 371,554 & 7.6\\
	\hline
	Total  & 260,190 & 3,605,693 & ——\\
	\hline
	\end{tabular}
	\begin{tablenotes}
        \footnotesize
        \item[*] The length represents the length of the utterance equal to the number of its words. 
      \end{tablenotes}
      \end{threeparttable}
\end{table}

\subsubsection{Jing Dong Dialogue Corpus}
The Customer Service Dataset, Jing Dong Dialogue Corpus (JDDC) \cite{chen2019jddc}, has been published as a part of JD contest\footnote{\url{http://jddc.jd.com/auth_environment}}. 
Analogously to the CDD dataset, the tested models will learn to generate the customer service employee's utterances based on the input represented by the prior part of the conversation.
The details of JDDC dataset are shown in Table \ref{tab:table2}.

\begin{table}[!t]
\small
    \caption{Statistics of Jing Dong Dialogue Corpus}
    \label{tab:table2}
    \vspace{-5pt}
    \begin{threeparttable}
    \setlength{\tabcolsep}{4.5mm}
	\begin{tabular}{c|c|c|c}
	
	\hline
	 ~ & Conversation & Utterance & length(avg) \\
	\hline
	Train & 261,282 & 3,135,377  & 9.4\\
	\hline
	Dev & 32,660  & 391,983 & 9.1\\
	\hline
	Test & 32,661  & 391,480 & 9.8\\
	\hline
	Total  & 326,603 & 3,918,840 &——\\
	\hline
	
	\end{tabular}
		\begin{tablenotes}
        \footnotesize
        \item[*] The length represents the length of the utterance equal to the number of its words. 
      \end{tablenotes}
      \end{threeparttable}
\end{table}

To motivate other scholars to investigate this novel and challenging problem we release our code as well as the datasets.

\subsection{Baselines}
To test our model we select several representative and state-of-the-art works in text generation as the baseline methods.

\begin{itemize}[leftmargin=*]

\item CNN-based models:
\begin{itemize}[leftmargin=*]
\item \textbf{ByteNet} \cite{kalchbrenner2016neural}: a one-dimensional Convolutional Neural Network.
\item \textbf{ConvS2S} \cite{gehring2017convolutional}: This approach uses the convolutional neural network as an encoder to solve the problem of long sequence training.
\end{itemize}

\item RNN-based models:
\begin{itemize}
\item  \textbf{LSTM} \cite{hochreiter1997long}: a unidirectional Long Short Term Memory network.
\item \textbf{S2S+attention} \cite{nallapati2016abstractive}: a model in which the encoder encodes the input sequence, while the decoder produces the target sequence. The attention mechanism is added to force the model to learn focus on specific parts of the input sequence when decoding.
\item \textbf{PGN} \cite{see2017get}: a commonly used method in the tasks of text generation and automatic summarization, which utilizes the copy mechanism in the decoder to effectively solve Out-Of-Vocabulary problem.
\end{itemize}

\item Transformer-based models:
\begin{itemize}
\item \textbf{Transformer} \cite{vaswani2017attention}: a neural network using positional encoding and multi-head self-attention mechanism.
\item \textbf{DAM} \cite{zhou2018multi}: a multi-turn conversation model which matches a response with its multi-turn context using dependency information based entirely on attention.
\item \textbf{ReCoSa} \cite{zhang2019recosa}: a multi-turn conversation model in which the self-attention mechanism is utilized to update both the context and masked response representation.

\item \textbf{Retrieval-guided} model \cite{cai2019retrieval}: a model in which the skeleton extraction is made by an interpretable matching model, and the following skeleton-guided response generation is accomplished by a separately trained generator.
\end{itemize}

\item Hierarchical LSTM-based models:
\begin{itemize}
\item \textbf{HRED} \cite{serban2015building}: a hierarchical RNN structure which enables to simultaneously model the sentence-layer information and the conversation-layer information in multi-turn conversation.
\item \textbf{EED} \cite{pandey2018exemplar}: a model which retrieves responses to create exemplar vectors and uses the vector to decode response.
\item \textbf{CCN}
\cite{ji-etal-2020-cross}: a model which uses the combination of copying the current context and the content from similar conversation.
\end{itemize}
\end{itemize}
It should be noted that the \textbf{Retrieval-guided} approach is a fusion of extractive and generative based models. In addition, the methods \textbf{EED} and \textbf{CCN} also utilize similar dialogues for modeling during training as in case of our model. 
It needs to be also pointed out that \textbf{CCN} and \textbf{EED} use multiple similar conversations during the training, and we chose the best result as the baseline; they are marked as \textbf{CCN}$_{best}$ and \textbf{EED}$_{best}$, respectively.

Furthermore, all the baselines were trained in the same way as the proposed model \textbf{DRMN} was trained to make fair comparison across the models.

\subsection{Evaluation Measures}
We adopt two evaluation methods to evaluate the performance of all the tested models.

\subsubsection{Automatic Evaluation}
Automatic evaluation adopts quantitative evaluation metrics commonly used in text generation tasks: BLEU \cite{papineni2002bleu} and ROUGE \cite{lin2004rouges}.
We regard BLEU and ROUGE scores as objective evaluation, serving as the measures of method performance \cite{song2016two}. 
We report ROUGE-1, ROUGE-L and BLEU to understand the performance of each model and their advantages as well as disadvantages.

\subsubsection{Human Evaluation}
In order to ensure the fluency and rationality of the generated utterances, we also qualitatively analyzed data through human evaluation.
We hired five well-educated NLP researchers to evaluate the quality of the generated utterances. We evaluated the effect of independent evaluation from two aspects: Relevance and Fluency \cite{ke2018generating,zhu2019retrieval}:

\begin{itemize}[leftmargin=*]
\item Relevance: the generated utterance is logically relevant to the conversation context and can provide meaningful information.
\item Fluency: Generated utterance is fluent and grammatical.
\end{itemize}
We randomly selected $300$ examples from the test set for each model.
For either aspect, we set three levels with scores: $+2$, $+1$, $0$, in which higher score stands for excellent. To compute the final scores from 5 annotators, we remove the highest score and the lowest score given by the annotators and then calculated the average of the remaining three scores. We report the average score and coefficient $\kappa$ which indicates the consistency of evaluation among annotators.

\begin{table}[!t]
\small
  \caption{Quantitative Evaluation. We report ROUGE-1 (R-1), ROUGE-L (R-L), and BLEU scores for each tested method. }
  \vspace{-8pt}
  \label{tab:result_discussion}
  \begin{center}
  \begin{tabular}{c|c|c|c|c|c|c}
    \hline
    \multicolumn{1}{c|}{\multirow{2}{*}{model}} &
    \multicolumn{3}{c|}{CDD} &
    \multicolumn{3}{c}{JDDC} \\ 
    \cline{2-7}
    & R-1 &  R-L & BLEU 
    & R-1 &  R-L & BLEU\\
    \hline
    ByteNet \cite{gehring2017convolutional}        
    &  33.68 & 32.99  &  16.91
    & 22.19  & 18.35  &  11.55 \\
    ConvS2S \cite{kalchbrenner2016neural}       
    & 35.92 & 31.48  & 16.34      
    & 26.53 & 21.08  & 11.64 \\
   
    \hline
    LSTM \cite{hochreiter1997long}           
    &  30.28 & 28.02  &  9.77 
    &  19.45 & 18.74  &  9.52  \\
    S2S+attention \cite{nallapati2016abstractive}  
    & 36.91 &  33.12  &  18.52
    & 28.44 &  22.34  &  13.42 \\
    PGN \cite{see2017get}            
    & 37.03  &  34.25  & 18.75   
    & 29.78  &  24.06  & 14.37 \\
    
    \hline
    Transformer \cite{vaswani2017attention}    
    & 37.59  &  34.93  & 18.58
    & 27.25  &  22.75  & 11.29 \\
    DAM \cite{zhou2018multi}
    & 38.28  & 35.27  &  20.83     
    & 28.86  & 23.79  &  13.95 \\
    ReCoSa \cite{zhang2019recosa}
    & 38.53  & 35.38  &  20.95     
    & 30.83  & 24.67  &  14.94 \\
    Retrieval-guided \cite{cai2019retrieval}
    & 37.27  & 34.75  &  19.26     
    & 28.75  & 22.28  &  12.89 \\

    \hline
    HRED \cite{serban2015building}
    & 38.22  & 35.74  &  20.71     
    & 28.01  & 23.28  &  13.86 \\
    EED$_{best}$ \cite{pandey2018exemplar}
    & 39.28  & 37.55  &  22.43      
    & 32.18  & 30.07  &  18.11\\
    CCN$_{best}$ \cite{ji-etal-2020-cross}
    & 41.10  & 39.82  & 24.75       
    & 34.17  & 32.37  & 19.53  \\
    
    \bottomrule
    DRMN$_{top-1}$
    & 43.79 & 39.23 & 23.11    	
    & 35.98 & 32.71 & 22.08 \\
    DRMN$_{top-2}$
    & 44.68  & 40.51 & 27.27    
    & $\textbf{36.31}$  & 33.19 & 23.37 \\
    DRMN$_{top-3}$
    & $\textbf{45.03}$  & $\textbf{43.09}$  &  $\textbf{28.96}$   
    & 36.15  & $\textbf{33.35}$  &  $\textbf{23.42}$ \\
    \hline
    \end{tabular}
    \vspace{-8pt}
\end{center} 
\end{table}

\subsection{Training Details}
For representing utterances, we set the dimensions of word embedding as $300$ and use word2vec to build the initial word vectors.
For the role information, the dimension of the role embedding is set to $100$ with random initialization.
In the encoder, the \textbf{DRMN} is implemented by two-layer LSTM networks with a hidden size of $300$. In this case, a combination of forward and backward LSTM gives us $600$ dimensions.
The dropout is set to $0.8$. 
Based on these settings, we optimize the objective function with the learning rate of $5e-4$. 
We perform the mini-batch gradient descent with a batch size of $32$.
During the experiment, we adopted cross-validation to ensure the rationality of the model. 


\begin{table}[!t]
\small
\centering
  \caption{Qualitative Evaluation. We report the average scores (Avg) and calculate the $\kappa$ values for relevance and fluency. 
  }
  \label{tab:result_discussion2}
  \setlength{\tabcolsep}{0.8mm}{
  \begin{tabular}[width=\linewidth]{c|c|c|c|c|c|c|c|c}
    \hline
    \multicolumn{1}{c|}{\multirow{3}{*}{model}} &
    \multicolumn{4}{c|}{CDD} & \multicolumn{4}{c}{JDDC} \\ 
    \cline{2-9}
    &\multicolumn{2}{c|}{Relevance} & \multicolumn{2}{c|}{Fluency}   &
    \multicolumn{2}{c|}{Relevance} & \multicolumn{2}{c}{Fluency}\\
     \cline{2-9}
    &  Avg & $\kappa$ &  Avg & $\kappa$ &  Avg & $\kappa$ &  Avg & $\kappa$  \\
    \hline
    ByteNet \cite{gehring2017convolutional}            
    & 0.63  & 0.62  & 1.01  & 0.71
    & 0.59  & 0.55 & 1.19   & 0.67\\
    ConvS2S \cite{kalchbrenner2016neural}    
    & 0.64  &  0.51 & 1.05  & 0.82
    & 0.67  &  0.71 & 1.13  & 0.56\\
    \hline
    
    LSTM \cite{hochreiter1997long}  
    & 0.54  & 0.48  & 0.93   & 0.61
    & 0.53  & 0.52  & 1.09   & 0.59\\
    S2S+attention \cite{nallapati2016abstractive}  
    & 0.89  & 0.55  & 1.32  & 0.69
    & 0.88 &  0.48 &  1.26  & 0.57\\
    PGN \cite{see2017get}      
    &  1.06 &  0.64 & 1.47  & 0.72
    &  0.96 &  0.69 & 1.52  & 0.53\\
    \hline
    Transformer \cite{vaswani2017attention}
    & 1.02  & 0.71  & 1.41  & 0.65
    & 0.83  & 0.56  & 1.42  & 0.73\\
    DAM \cite{zhou2018multi}
    & 1.04  & 0.77  & 1.47  & 0.57 
    & 0.88  & 0.58  & 1.51  & 0.68 \\
    ReCoSa \cite{zhang2019recosa}
    & 1.06  & 0.67  & 1.55  & 0.65 
    & 0.91  & 0.71  & 1.59  & 0.55 \\
    Retrieval-guided \cite{cai2019retrieval}
    & 0.96  & 0.72  & 1.43  & 0.59 
    & 0.92  & 0.64  & 1.48  & 0.63 \\
    \hline
    HRED \cite{serban2015building}
    & 1.01	& 0.49  & 1.43  & 0.62  
    & 0.73  & 0.71  & 1.47  & 0.54 \\
    EED$_{best}$ \cite{pandey2018exemplar}
    & 1.11  & 0.63  & 1.62  & 0.73 
    & 1.07  & 0.65  & 1.65  & 0.54 \\
    CCN$_{best}$ \cite{ji-etal-2020-cross}
    & 1.12  & 0.66  & 1.69  & 0.68 
    & 1.01  & 0.72  & $\textbf{1.77}$  & 0.70 \\
    
    \bottomrule
    DRMN$_{top-1}$
    & 1.13 & 0.75 & 1.68  & 0.74 
    & 1.01 & 0.64 & 1.69  & 0.79 \\
    DRMN$_{top-2}$
    & 1.12 & 0.64 & 1.71  & 0.69  
    & 1.05 & 0.68 & 1.73  & 0.65 \\
    DRMN$_{top-3}$
    &  $\textbf{1.15}$ & 0.62 & $\textbf{1.74}$  & 0.62  
    & $\textbf{1.09}$ & 0.67 & 1.72  & 0.63  \\
  \hline
    \end{tabular}}
    \vspace{-8pt}
\end{table}

%% file: 5_Result_discussion.tex
\vspace{-4pt}
\section{Experimental Results}

\subsection{Overall Performance}
In this section we conduct the analysis of results from diverse perspectives to thoroughly evaluate the performance of the proposed model. One thing to note before delving into details is that to prove the impact of ESM module in our \textbf{DRMN} model, we applied different numbers of similar conversations, i.e., utilizing the most similar conversation (top-1), the top two similar ones (top-2), and the top three similar ones (top-3).
  

\textbf{Quantitative Comparison against baselines.}
The quantitative performance of all the tested methods is reported in Table \ref{tab:result_discussion}. As Table \ref{tab:result_discussion} shows, the proposed approach \textbf{DRMN}$_{top-3}$ significantly outperforms all the baselines in ROUGE and BLEU metrics over the two datasets.

Compared with the first three groups of baselines (CNN-based, RNN-based and Transformer-based), \textbf{DRMN} with its variants perform significantly better than these models by a large margin. The group of Hierarchical LSTM-based models shows better performance than the the first three groups. 
It demonstrates the effectiveness of the hierarchical infrastructure for modeling conversations by capturing word-level and sentence-level dependencies, which can further improve the quality of the generated text. 

Moreover, we notice the higher performance of the two baseline methods \textbf{EED}$_{best}$ and \textbf{CCN}$_{best}$ compared to the other baselines. It indicates the advantage of leveraging similar conversations in the task of text generation. Compared with the infrastructures of these two best baselines, \textbf{DRMN} can gradually memorize keywords, phrases, and sentences from similar dialogues to assist the generation of the target conversation. \textbf{CCN} highly relies on the copy mechanism on the decoder side, which tends to cause the problem of copy position error during the generation process, as well as its limitation in copying the keywords from remote sentences. On the other hand, \textbf{EED} uses only similar dialogues as a knowledge-assisted generation, and cannot extract key words, phrases, and sentences from similar dialogues.
 
\textbf{Qualitative Comparison against baselines.}
The quantitative performance of all the tested methods is reported in Table \ref{tab:result_discussion2}. To be fair, for each input, we shuffled the output generated by all the models and then let the annotators evaluate them. 
As noted earlier, $\kappa$ indicates the consistency of the annotator's evaluation. 
The observed $\kappa$ coefficient values that range between 0.48 and 0.82 indicate middle and upper agreement.
We found that the relevance and fluency compared to the best performing baseline model (CCN) increased by $2.7\%$ and $2.9\%$ in the CDD, respectively. For the JDDC dataset, although the fluency of our qualitative evaluation is lower than that of CCN dataset, the relevance was still improved by $7.9\%$.


\textbf{The impact of the number of similar conversations used.}
We observe the increasing performance as the number of referred similar conversations increases (see the results of \textbf{DRMN}$_{top-1}$, \textbf{DRMN}$_{top-2}$, \textbf{DRMN}$_{top-3}$ in Table \ref{tab:result_discussion}). 
As mentioned above, the prediction of the model appears to improve along with the increase in the number of similar conversations, indicating that similar conversations play an important role in reading and memory.
However, the increase of the number of similar conversations also adds a certain degree of complexity to train the model. It makes the time cost of model training higher as well as results in larger space cost. In order to balance the effectiveness and the training cost, we choose at most three similar conversations (top-3) for experiments to verify the validity of our model. 

\textbf{Comparison based on datasets.}
In addition, we compare the results obtained for the two datasets. 
We observe that the results on the CDD dataset are better than the ones on JDDC both in the quantitative and qualitative evaluations. 
After manually investigating the contents of our datasets,
we concluded the following three possible reasons.
First, compared with the court debate scenario, customer service conversations are much more open due to a large number of types and aspects of the commodities. It makes the task of text generation more difficult. 
Second, the utterances of customers and customer service staff tend to be more colloquial, while the judge's utterances are more strict and formal. We note that colloquial sentences may cause difficulties for language model training due to a large variety of non-standard expressions. 
Last, in the JDDC dataset, there is higher number of the same or similar utterances. For example, the phrases like "Welcome back again!", "Can I help you?" appear quite repetitively.

\begin{table}[!t]
\small
  \caption{Ablation study: quantitative evaluation.}
  \vspace{-8pt}
  \label{tab:abtest}
  \begin{center}
  \begin{tabular}{c|c|c|c|c|c|c}
    \hline
    \multicolumn{1}{c|}{\multirow{2}{*}{model}} &
    \multicolumn{3}{c|}{CDD} &
    \multicolumn{3}{c}{JDDC} \\ 
    \cline{2-7}
    & R-1 &  R-L & BLEU 
    & R-1 &  R-L & BLEU\\
     \hline
    TC+SC
    & 38.71  & 37.09  &  21.86     
    & 31.75  & 24.92  &  15.65 \\
    -ESM
    & 37.53  & 35.29  &  19.02     
    & 29.06  & 23.72  &  13.74 \\
    \hline
    DRMN$_{top-1}$
    & 43.79 & 39.23 & 23.11    	
    & 35.98 & 32.71 & 22.08 \\
    DRMN$_{top-2}$
    & 44.68  & 40.51 & 27.27    
    & 36.31  & 33.19 & 23.37 \\
    DRMN$_{top-3}$
    & $\textbf{45.03}$  & $\textbf{43.09}$  &  $\textbf{28.96}$   
    & $\textbf{36.15}$  & $\textbf{33.35}$  &  $\textbf{23.42}$ \\
  \hline
    \end{tabular}
    \vspace{-8pt}
\end{center} 
\end{table}

\begin{table}[!t]
\small
\centering
  \caption{Ablation study: qualitative evaluation.}
  \vspace{-8pt}
  \label{tab:abtest2}
  \begin{tabular}[width=\linewidth]{c|c|c|c|c|c|c|c|c}
    \hline
    \multicolumn{1}{c|}{\multirow{3}{*}{model}} &
    \multicolumn{4}{c|}{CDD} & \multicolumn{4}{c}{JDDC} \\ 
    \cline{2-9}
    &\multicolumn{2}{c|}{Relevance} & \multicolumn{2}{c|}{Fluency}   &
    \multicolumn{2}{c|}{Relevance} & \multicolumn{2}{c}{Fluency}\\
     \cline{2-9}
    &  Avg & $\kappa$ &  Avg & $\kappa$ &  Avg & $\kappa$ &  Avg & $\kappa$  \\
    \hline
    TC+SC
    & 1.09  & 0.81  & 1.63  & 0.48 
    & 0.96  & 0.61  & 1.66  & 0.62 \\
    -ESM
    & 1.03	& 0.58  & 1.51  & 0.71  
    & 0.79  & 0.53  & 1.49  & 0.66 \\
    \hline
    DRMN$_{top-1}$
    & 1.13 & 0.75 & 1.68  & 0.74 
    & 1.01 & 0.64 & 1.69  & 0.79 \\
    DRMN$_{top-2}$
    & 1.12 & 0.64 & 1.71  & 0.69  
    & 1.05 & 0.68 & $\textbf{1.73}$  & 0.65 \\
    DRMN$_{top-3}$
    &  $\textbf{1.15}$ & 0.62 & $\textbf{1.74}$  & 0.62  
    & $\textbf{1.09}$ & 0.67 & 1.72  & 0.63  \\
  \hline
    \end{tabular}
    \vspace{-8pt}
\end{table}

\subsection{Ablation test}
To assess the contribution of ESM module and similar conversations, we next conduct the ablation tests. 
To prove the effectiveness of ESM module, we remove it from the \textbf{DRMN} model (removing the entire second part in Figure \ref{fig:network}), denoted as \textbf{-ESM}. To justify our way of integrating the similar conversations with the target one, we simply concatenate them together as input (denoted as \textbf{TC+SC}), rather than modeling them in an interactive way as we have done for \textbf{DRMN}. 
Tables \ref{tab:abtest} and Table \ref{tab:abtest2} report the evaluation scores in terms of the quantitative and qualitative analysis, respectively.

According to the results shown in Tables \ref{tab:abtest} and Table \ref{tab:abtest2}, we notice a dramatic decrease in the performance of \textbf{-ESM} (decrease of $34.3\%$ on CDD and $41.3\%$ on JDDC, as measured by BLEU score). Similarly, the variant \textbf{TC+SC} has also experienced a large decrease in performance but less than that in \textbf{-ESM}. It shows that similar dialogues still play a certain role, but compared with the way of interactive modeling in \textbf{DRMN}, a straightforward concatenation \textbf{TC+SC} has limited effect. This confirms the effectiveness of the proposed circular reading and memory module. 

\begin{figure*}
    \centering
    \includegraphics[width=1\textwidth]{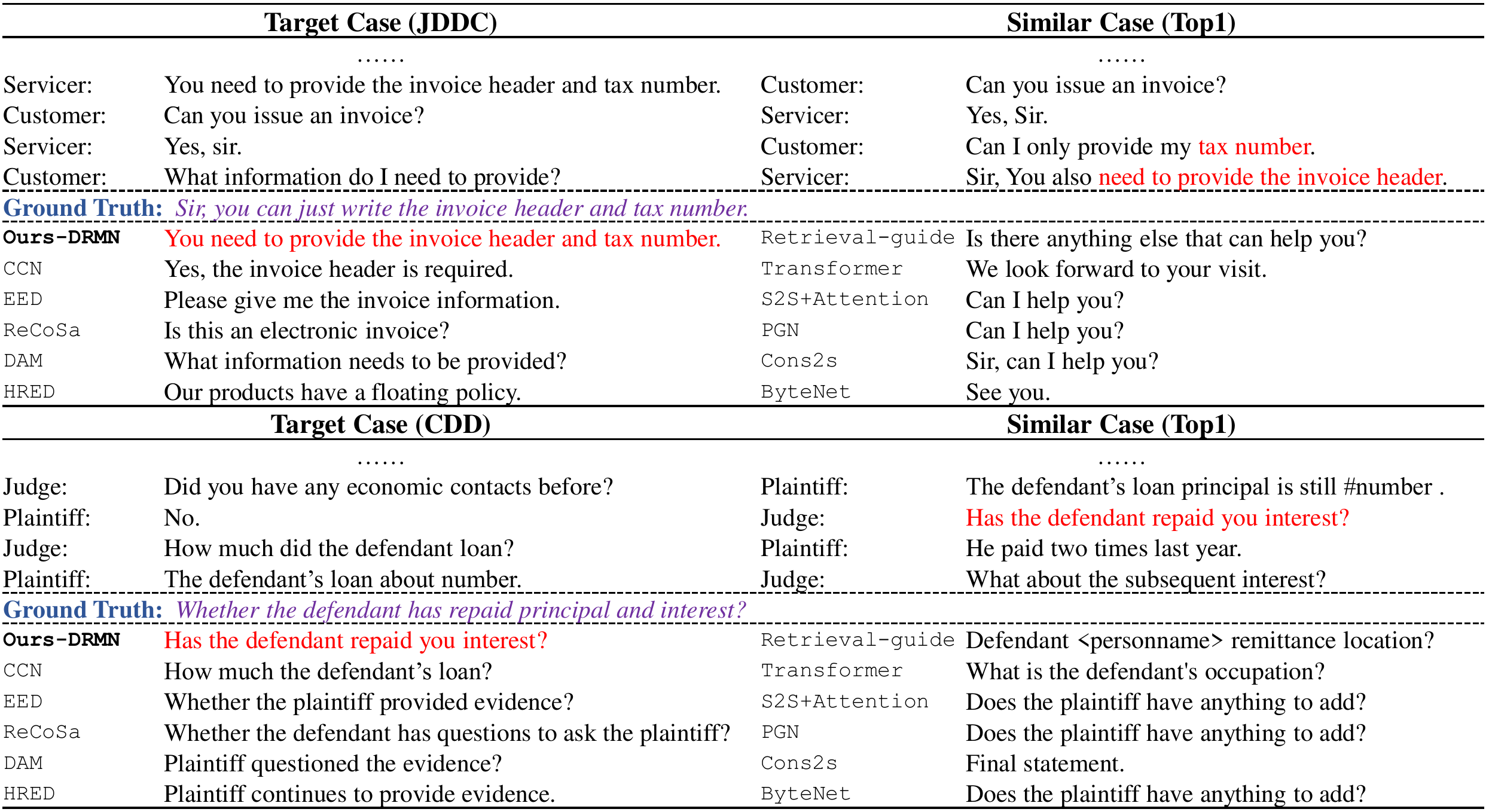}
    \caption{Case Study. We take two examples (target cases shown on the left side) from the judicial data and customer service data. We then show the following ground truth utterance as well as the utterances generated by different models. In addition, similar cases are displayed on the right. DRMN can memorize similar conversation information and accurately locate related entity, phrases, as well as sentences in similar conversations through the current conversation logic (the red text represents generated utterances that also appeared in similar conversations, and which our model can accurately remember.).}
    \label{fig:study}
\end{figure*}

\begin{figure}
    \centering
    \includegraphics[width=\columnwidth]{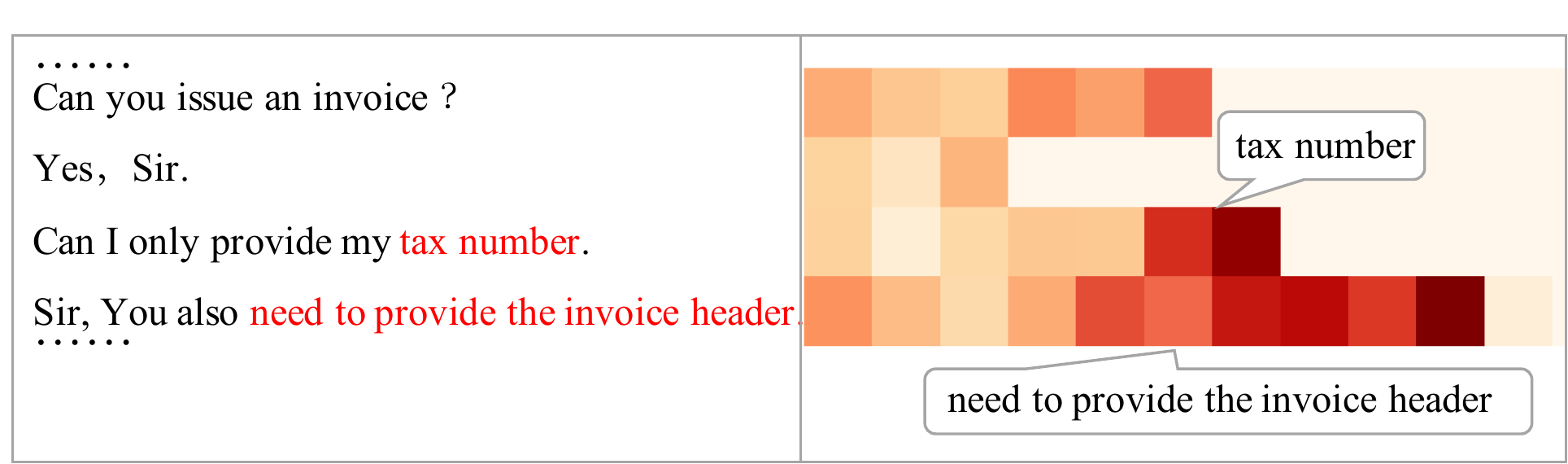}
    \vspace{-4pt}
    \caption{Visual analysis graph: the diagram on the right shows the significance of the DRMN model for memorizing words in similar conversations based on the sub-part of the example shown in Figure \ref{fig:study}; the depth of the color represents the importance of words and the darker the color, the greater the weight of the word.}
    \vspace{-6pt}
    \label{fig:visual}
\end{figure}

\subsection{Case study}
To help with better understanding of our model's performance, we demonstrate two cases in Figure \ref{fig:study}. The figure shows the results obtained by different models such that the left side represents the target conversation, and the right side represents its similar conversation in the dataset. We show the ground truth utterance (in purple color) as well as the utterances generated by different models. We also highlight the utterance delivered by our model and the relevant context in the similar case (in red color). 
We can first observe that ESM can be extracted either from a single sentence (such as the CDD example in Figure \ref{fig:study}, for which the ESM comes from the third last sentence of a similar conversation), or it can originate from multiple sentences (such as the JDDC example in Figure \ref{fig:study}, in which the ESM comes from the last two sentences of the similar conversation).

As shown in the Figure \ref{fig:study}, compared with the best performing baselines, especially \textbf{CCN} and \textbf{EED}, our model can better capture important entities, phrases, as well as the sentences from similar conversations.
For example, "tax number" and "need to provide the invoice header" are recalled from similar conversations.
Since the memory module in \textbf{DRMN} uses a self-attention mechanism, parallel calculations can be performed for long sentences. Hence the long-term memory can be achieved.

We next verify whether the performance improvements are obtained thanks to the detected relevant similar conversations.
We analyze the attention weights ($Y_L$ in Figure \ref{fig:network}) of the similar conversations for the first example of our case study. As shown in Figure \ref{fig:visual}, the darker the color, the higher the weight of the word is, and the greater is the impact on the context (it means these words have higher importance). We can observe that \textbf{DRMN} selects keywords by assigning them high weights; these words are accurately memorized.

\subsection{Error analysis}
To explore the limitations of our model, we also analyze the generated utterances, summarize the problems that occur, and explore the optimization solutions.

After conducting statistical analysis, we found that \textbf{DRMN} performs worse for low-frequency utterances/keywords or ones that do not appear in the target conversation or similar conversation.
In particular, there are $43\%$ errors\footnote{The error refers here to the generated text for which either the relevance or the fluency score equals 0.} belonging to this case in the JDDC dataset.
For example, in the sentence \textit{"Your order was successfully intercepted"}, the word \textit{"intercepted"} is a low-frequency word, and it has not appeared in the target conversation neither in the similar conversation. 
Similarly, in the CDD dataset, such a problem caused $45\%$ of errors, e.g., in sentences like: \textit{"Why was the application for investigation and evidence collection not submitted until today?"}, \textit{"application"} and \textit{"collection"} are in low-frequency in legal trial scenario.
In addition, $42\%$ of errors in the JDDC dataset occur when specific attributes of products are mentioned which tend to appear sparsely in the dataset, e.g., \textit{"The hard disk capacity of this computer is 500G, and the memory is 16G"}. 
Finally, it is worth mentioning that the proposed model has worse performance when generating long sentences. 

Table \ref{tab:error_analysis} shows the statistics of the cases with fluency score equal to $0$ for two best performing baselines and our proposed model. We can observe that all the tested models face significant difficulties in generating long sentences. Among all the utterances with a fluency score equal to $0$, the proportion of long utterances\footnote{I.e. the utterances with length greater than $10$.} for \textbf{DRMN} takes up to $78.6\%$ ($85\%$ for \textbf{CCN} and $92\%$ for \textbf{EED} ). It proves the superiority of \textbf{DRMN} even in very hard cases.

The improvement of the pre-training language models and constructing retrieval database could be promising approaches for future research to address the above-mentioned problems.

\begin{table}
\small
    \caption{Error analysis for the tested models on the cases with fluency score equal to 0.}
    \vspace{-5pt}
    \label{tab:error_analysis}
	\begin{tabular}{c|c|c|c}
	\hline
	 Model & \textbf{DRMN} & \textbf{CCN }& \textbf{EED} \\
	 \hline
	\#long utterance/ratio & 11/78.6\% & 17/85\%  & 23/92\%\\
	\hline
	\#short utterance/ratio & 3/21.4\%  & 3/15\% & 2/8\%\\
	\hline
	\end{tabular}
\vspace{-5pt}
\end{table}

%% file: 6_Conclusion_outlook.tex
\vspace{-12pt}
\section{Conclusion and Future Work}
The motivation behind our work is to improve the efficiency of conversation generation in specific domains. In particular, we propose a novel neural network structure called Deep Reading Memory Networks (DRMN), which enhances the expression of the model by reading and memorizing similar dialogues, as well as improving the quality of the generated text. Unlike the prior research, the proposed approach does not need to leverage any external knowledge, thus maintaining high field adaptability.
We conduct experiments on two different datasets with both quantitative and human evaluation to validate the effectiveness of our proposed model. 
Experimental results indicate DRMN's superiority when compared with a number of existing state-of-the-art text generation models, which suggests that the Deep Reading Memory Networks can successfully improve the conversation generation performance.

In the future, we will further investigate other content generation problems by leveraging multi-granularity memorizing and copying mechanism. The current study serves as the methodological foundation for this goal. 

\section*{Acknowledgments}
This work has been supported by the National Key R\&D Program of China (2018YFC0830200; 2018YFC0830206; 2018YFC0830700 and 2020YFC0832505).